\pdfoutput=1

\documentclass[11pt]{article}
\usepackage{tikz}
\usetikzlibrary{shapes.geometric, arrows}
\usepackage{amsmath}

\usepackage[preprint]{acl}

\usepackage{times}
\usepackage{latexsym}
\usepackage{algorithm}
\usepackage{algorithmic}

\usepackage[T1]{fontenc}

\usepackage[utf8]{inputenc}

\usepackage{microtype}

\usepackage{inconsolata}

\usepackage{graphicx}

 \title{Is your LLM trapped in a Mental Set?\\Investigative study on how mental sets affect the reasoning capabilities of LLMs
}

\author{
        Saiful Haq\textsuperscript{1,2}, \textbf{Niyati Chhaya\textsuperscript{2}}, \textbf{Piyush Pandey\textsuperscript{1}}, \textbf{Pushpak Bhattacharya\textsuperscript{1}}\\ 
        IIT Bombay\textsuperscript{1}, Hyperbots Inc\textsuperscript{2}\\
        \texttt{saifulhaq@cse.iitb.ac.in}
        }

\begin{document}
\maketitle
\begin{abstract}
In this paper, we present an investigative study on how \textbf{Mental Sets} influence the reasoning capabilities of LLMs. LLMs have excelled in diverse natural language processing (NLP) tasks, driven by advancements in parameter-efficient fine-tuning (PEFT) and emergent capabilities like in-context learning (ICL). For complex reasoning tasks, selecting the right model for PEFT or ICL is critical, often relying on scores on benchmarks such as MMLU, MATH, and GSM8K. However, current evaluation methods, based on metrics like F1 Score or reasoning chain assessments by larger models, overlook a key dimension: adaptability to unfamiliar situations and overcoming entrenched thinking patterns. In cognitive psychology, Mental Set refers to the tendency to persist with previously successful strategies, even when they become inefficient - a challenge for problem solving and reasoning. We compare the performance of LLM models like Llama-3.1-8B-Instruct, Llama-3.1-70B-Instruct and GPT-4o in the presence of mental sets. To the best of our knowledge, this is the first study to integrate cognitive psychology concepts into the evaluation of LLMs for complex reasoning tasks, providing deeper insights into their adaptability and problem-solving efficacy.
\end{abstract}

\section{Introduction}
Recent advancements in transformer architectures \cite{vaswani2017attention} have profoundly reshaped NLP and vision-language tasks. Models such as LLaMA-3 \cite{dubey2024llama}, Phi-4 \cite{abdin2024phi}, Mixtral \cite{jiang2024mixtral}, Deepseek-v3 \cite{liu2024deepseek}, InternVL-2.5 \cite{chen2024expanding}, and Pixtral \cite{agrawal2024pixtral} have set new benchmarks in reasoning and generalization capabilities. These open-source large language models (LLMs) and vision-language models (VLMs) have demonstrated remarkable performance on challenging benchmarks like MMLU \cite{hendrycks2020measuring}, GSM8K \cite{cobbe2021gsm8k}, and MATH \cite{hendrycks2021measuring}, solidifying their role in solving reasoning-intensive tasks.  

One of the most significant developments driving these advancements is in-context learning (ICL) \cite{dong2022survey}. Unlike traditional training paradigms that require fine-tuning on task-specific data, ICL enables models to adapt to new tasks through examples embedded directly in the input prompts. This paradigm not only enhances task adaptability but also reduces computational overhead. The flexibility and scalability of ICL have propelled its adoption in solving problems across diverse domains, from mathematical reasoning to commonsense understanding.  

Despite their success, traditional evaluation metrics such as accuracy, F1 score, and reasoning chain assessments primarily capture task performance in well-defined or familiar scenarios. These benchmarks, however, fail to account for an essential aspect of reasoning: adaptability in the face of novelty and the ability to overcome entrenched problem-solving patterns. This dimension aligns with the concept of a \textbf{Mental Set} \cite{ollinger2008investigating} in cognitive psychology—a phenomenon where reliance on previously successful strategies inhibits the adoption of more efficient solutions in novel situations. Mental set, often described as cognitive rigidity, emerges when repeated success with specific approaches reinforces their use, even in cases where such approaches are suboptimal or ineffective. While this tendency streamlines problem-solving in routine contexts, it presents a critical bottleneck when models encounter unfamiliar challenges that demand innovative solutions.

In contrast to mental set, \textbf{Insight} represents the ability to overcome these entrenched patterns through a reorganization of mental representations, enabling novel solutions to emerge. Insightful problem-solving is characterized by sudden breakthroughs, often accompanied by the well-documented "aha!" moment. This process is inherently different from systematic problem-solving as it involves either decomposing familiar cognitive chunks or relaxing implicit constraints that limit how a problem is conceptualized. For instance, Duncker’s candle problem \cite{duncker1945problem} and Luchins' water jug experiments \cite{luchins1959rigidity} demonstrate how initial problem representations can either obstruct or facilitate insight, depending on whether solvers can reframe their approach. 

The interplay between mental set and insight is particularly relevant when assessing reasoning capabilities, as models that exhibit high adaptability must strike a balance between leveraging prior knowledge and restructuring it when necessary. Understanding how these cognitive phenomena interact provides deeper insight into the flexibility and robustness of AI systems when faced with complex, unfamiliar tasks. Our contribution is:
\begin{itemize}
\item An investigative study on how \textbf{Mental Sets} influence the reasoning capabilities of LLMs (Llama-3.1-8B-Instruct, Llama-3.1-70B-Instruct and GPT-4o). To the best of our knowledge, this is the first study to integrate cognitive psychology concepts into the evaluation of LLMs for complex reasoning tasks, providing deeper insights into their adaptability and problem-solving efficacy.

\end{itemize}

\section{Related work}
Mental set \cite{ollinger2008investigating} occurs when individuals continue to use a familiar strategy, even when it is no longer the most efficient approach. This is evident in problem-solving contexts like the Water-Jug problem \cite{luchins1959rigidity}, where participants persist in applying a complex method despite the availability of a simpler solution. Such entrenched thinking patterns can limit problem-solving effectiveness. For example, students might repeatedly use the "add-subtract" method in solving problems like 5 + 2 + 3 = ? + 3, even when the "grouping" method would be faster and less cognitively demanding. This rigidity in thinking is a clear example of mental set interfering with problem-solving flexibility. On the other hand, procedural flexibility \cite{decaro2016inducing} refers to the ability to shift between various strategies or methods when solving problems, particularly in mathematics, where students often learn multiple approaches to tackle different types of problems. This flexibility enables them to select the most efficient method depending on the problem's context. Research has shown that procedural flexibility is linked to a deeper conceptual understanding. For instance, when students are exposed to different problem-solving strategies, they become better equipped to adapt their approach to novel or unfamiliar challenges. Educational psychology studies suggest that students with greater procedural flexibility tend to experience reduced cognitive load and improved accuracy in their problem-solving. For example, in mathematics, students may be taught multiple strategies, such as the "add-subtract" method and the "grouping" method. The latter, which simplifies cognitive demands, may be more efficient in some cases. Flexible students are more likely to switch between methods depending on the problem, enhancing their understanding of key concepts like mathematical equivalence. While procedural flexibility is a valuable skill, mental set can hinder its development.








\section{Dataset}
The dataset used in this study is based on the problems outlined in \cite{decaro2016inducing}, as summarized in Table \ref{tab:problems}. It consists of two types of mathematical equivalence problems: complex problems and shortcut problems. Each type includes six problems.
\begin{itemize}
    \item Complex problems require multi-step strategies to solve. These problems involve various addition operations where multiple numbers must be added before solving for the missing value.
    \item Shortcut problems are designed for quicker resolution by leveraging repeated addends on both sides of the equation, enabling simpler and more efficient solutions.
\end{itemize}
The numbers used in both problem sets are consistent to ensure that any differences in problem-solving difficulty arise solely from the required strategy, rather than the numerical values themselves. The dataset is structured to evaluate how the order of problem presentation influences participants' (or LLMs') use of efficient problem-solving strategies.

\section{Experiment Setup}
\begin{table*}[h!]
  \centering
  \begin{tabular}{llcccc}
    \hline
    \textbf{Model} & \textbf{Metric} & \textbf{SP} & \textbf{SP+CPF} & \textbf{CP} & \textbf{CP+SPF} \\ 
    \hline
    \verb|Llama-3.1-8b-instruct| & EM & 0.16 & 0 & 0 & 0 \\ 
    & Steps & 1 & x & x & x \\ 
    \verb|Llama-3.1-70b-instruct| & EM & 0 & 0.16 & 0 & 0.66 \\ 
    & Steps & x & 1 & x & 1 \\ 
    \verb|GPT-4o| & EM & 0.33 & 0.66 & 0.66 & 0.83 \\ 
    & Steps & 1 & 1 & 1 & 1 \\ 
    \hline
  \end{tabular}
\caption{Performance comparison of models on shortcut problems (Sh) and complex problems (CP) across different conditions with few-shot prompting. The conditions include SP+CPF (shortcut problems with complex problems first) and CP+SPF (complex problems with shortcut problems first), along with their respective baselines (SP and CP). The metrics reported are Exact Match (EM) scores, averaged over six input problems per condition, and the number of steps required to reach the solution, averaged over the input problems where the model successfully arrives at the correct solution.}
  \label{tab:merged_results_1}
\end{table*}
\begin{table*}[h!]
  \centering
  \begin{tabular}{llcccc}
    \hline
    \textbf{Model} & \textbf{Metric} & \textbf{SP} & \textbf{SP+CPF} & \textbf{CP} & \textbf{CP+SPF} \\ 
    \hline
    \verb|Llama-3.1-8b-instruct| & EM & 0.5 & 0.66 & 0.33 & 0.83 \\ 
    & Steps & 3 & 3 & 3 & 3 \\ 
    \verb|Llama-3.1-70b-instruct| & EM & 0.5 & 0.66 & 0.66 & 0.83 \\ 
    & Steps & 3 & 3 & 3 & 3 \\ 
    \verb|GPT-4o| & EM & 0.66 & 0.66 & 0.83 & 0.83 \\ 
    & Steps & 3 & 3 & 3 & 3 \\ 
    \hline
  \end{tabular}
\caption{Performance comparison of models on shortcut problems (Sh) and complex problems (CP) across different conditions with fewshot prompting + zeroshot chain of thought. The conditions include SP+CPF (shortcut problems with complex problems first) and CP+SPF (complex problems with shortcut problems first), along with their respective baselines (SP and CP). The metrics reported are Exact Match (EM) scores, averaged over six input problems per condition, and the number of steps required to reach the solution, averaged over the input problems where the model successfully arrives at the correct solution.}
  \label{tab:merged_results_2}
\end{table*}
The experimental setup is designed to evaluate the ability of large language models (LLMs) to solve mathematical equivalence problems, with a focus on the efficiency of strategy use (complex vs. shortcut strategies) based on the order of problem presentation. The experiment is adapted from the methodology proposed by \cite{decaro2016inducing} and consists of two conditions:
\begin{itemize}
    \item \textbf{Complex-first condition:} LLMs solve six complex problems first, followed by six shortcut problems.
    \item \textbf{Shortcut-first condition:} LLMs solve six shortcut problems first, followed by six complex problems.
\end{itemize}
Each LLM is presented with the same set of problems, encompassing both complex and shortcut problems. For each problem, the LLMs are tasked with solving the equation by filling in the missing number, demonstrating their reasoning, and applying the most efficient strategy. The LLMs evaluated in this experiment include: Llama-3.1-8B-Instruct, Llama-3.1-70B-Instruct and GPT-4o. We employ both few-shot prompting and few-shot prompting with zero-shot chain-of-thought reasoning \cite{wei2022chain}.

The evaluation metrics used are as follows: Exact Match, which measures whether the model produces the correct final answer and Steps, which tracks the number of steps taken to arrive at the correct solution.



\begin{table}[h!]
  \centering
  \begin{tabular}{llc}
    \hline
    \textbf{Problem Type} & \textbf{Input} & \textbf{Output} \\ 
    \hline
    \verb|complex| & \verb|7+5+9=3+?| & \verb|18| \\ 
    \verb|complex| & \verb|4+14+8=6+?| & \verb|20| \\ 
    \verb|complex| & \verb|15+3+9=13+?| & \verb|11| \\ 
    \verb|complex| & \verb|9+7+6=?+3| & \verb|19| \\ 
    \verb|complex| & \verb|14+5+3=?+7| & \verb|15| \\ 
    \verb|complex| & \verb|6+3+12=?+15| & \verb|6| \\ 
    \verb|shortcut| & \verb|7+5+9=7+?| & \verb|14| \\ 
    \verb|shortcut| & \verb|4+14+8=4+?| & \verb|22| \\ 
    \verb|shortcut| & \verb|15+3+9=15+?| & \verb|12| \\ 
    \verb|shortcut| & \verb|9+7+6=?+6| & \verb|15| \\ 
    \verb|shortcut| & \verb|14+5+3=?+3| & \verb|15| \\ 
    \verb|shortcut| & \verb|6+3+12=?+12| & \verb|9| \\ 
    \hline
  \end{tabular}
  \caption{Complex and Shortcut Problems}
  \label{tab:problems}
\end{table}


\section{Results}
Performance, as measured by EM score, improves for all LLM models when using Few-shot+Zero-shot Chain-of-Thought prompting (FS+CoT), while they perform poorly with standard few-shot prompting (FS) alone. Incorporating in-context examples, whether under the CPF or SPG conditions, boosts EM scores, except for the SP+CPF condition with the Llama-3.1-8b-instruct model using FS prompting. However, these in-context examples do not influence the number of steps needed to reach the final solution. For successful cases, FS prompting generally requires fewer steps than FS+CoT. Notably, under FS+CoT, Llama-3.1-70b-instruct and GPT-4o show better performance on complex problems (CP) compared to shortcut problems (SP) in terms of EM score.

Humans typically require just one step to solve problems in SP and SP+CPF, and LLMs demonstrate this capability in a limited number of successful cases, as shown in Table \ref{tab:merged_results_1}. However, when using CoT prompting to improve performance, LLMs tend to take more than one step to solve problems, as illustrated in Table \ref{tab:merged_results_2}, reflecting a mental set shift induced by CoT reasoning.





\section{Summary, Conclusion and Future Work}
We present an investigative study on how mental sets influence the reasoning capabilities of LLMs. Inspired by cognitive psychology, we set up an experiment to compare the problem-solving abilities of three LLM models—Llama-3.1-8B-Instruct, Llama-3.1-70B-Instruct, and GPT-4o—on a mathematical equivalence dataset. Our findings show that in-context examples, whether under the CPF or SPF conditions, enhance performance but do not affect the number of steps taken to reach the final solution. Interestingly, while Few-shot (FS) prompting typically requires fewer steps, Few-shot+Zero-shot Chain-of-Thought (FS+CoT) prompting leads to more complex reasoning, as indicated by the increase in the number of steps required to solve problems. These results suggest that while humans can solve simpler problems in a single step, LLMs require more than one step to achieve higher accuracy, reflecting presence of mental sets. Future work includes expanding this dataset to test problem solving in VLMs.

\bibliography{main}




\end{document}